\documentclass[11pt, svgnames]{article}

\usepackage[preprint]{tmlr}

\usepackage[utf8]{inputenc}

\usepackage{ amsmath }
\usepackage{ amssymb }
\usepackage{booktabs}
\usepackage{multirow}
\usepackage{graphicx}
\usepackage{hyperref}

\hypersetup{
     colorlinks=true,
     linkcolor=blue,
     filecolor=blue,
     citecolor = blue,      
     }
     
\usepackage{natbib}
\title{Synthcity: facilitating innovative use cases of synthetic data in different data modalities}

\author{
      \name Zhoazhi Qian \email zhaozhi.qian@maths.cam.ac.uk\\
      \addr DAMTP, University of Cambridge
      \AND
      \name Bogdan-Constantin Cebere \email bcc38@cam.ac.uk \\
      \addr DAMTP, University of Cambridge
      \AND
      \name Mihaela van der Schaar \email mv472@cam.ac.uk\\
      \addr DAMTP, University of Cambridge\\
      The Alan Turing Institute}

\begin{document}

\maketitle

\begin{abstract}
    Synthcity is an open-source software package for innovative use cases of synthetic data in ML fairness, privacy and augmentation across diverse tabular data modalities, including static data, regular and irregular time series, data with censoring, multi-source data, composite data, and more. Synthcity provides the practitioners with a single access point to cutting edge research and tools in synthetic data. It also offers the community a playground for rapid experimentation and prototyping, a one-stop-shop for SOTA benchmarks, and an opportunity for extending research impact. The library can be accessed on \href{https://github.com/vanderschaarlab/synthcity}{GitHub} and \href{https://pypi.org/project/synthcity/}{pip}. We warmly invite the community to join the development effort by providing feedback, reporting bugs, and contributing code. 
\end{abstract}

\section{Synthetic data technology promises to empower AI}
\label{sec:Introduction}

Access to high quality data is the lifeblood of AI. Although AI holds strong promise in numerous high-stakes domains, the lack of high-quality datasets creates a significant hurdle for the development of AI, leading to missed opportunities. 
Specifically, three prominent issues contribute to this challenge: \textit{data scarcity}, \textit{privacy}, and \textit{bias}  \citep{mehrabi2021survey, gianfrancesco2018potential,tashea_17AD,dastin}. As a result, the dataset may not be available, accessible, or suitable for building performant and socially responsible AI systems \citep{sambasivan2021everyone}.

Synthetic data has the potential to fuel the development of AI by unleashing the information in datasets that are small, sensitive or biased. This topic has recently achieved much excitement and attention in the AI community, which led to the proposal of many novel methodologies \citep{jordon2018pate,yoon2020anonymization,ho2021dp, mehrabi2021survey, van2021decaf, zhu2017unpaired,yoon2018radialgan,saxena2021generative}. In contrast to traditional generative models, whose sole objective is to learn the distribution of the input data, the new class of generators are designed to produce high-fidelity synthetic data while satisfying additional constraints and properties, such as differential privacy \citep{dwork2008differential}, counterfactual fairness \citep{kusner2017counterfactual}, and causal invariances \citep{pearl2009causality}. 
Training downstream AI algorithms on the synthetic data would imbue them with these desirable properties, thereby providing better privacy, fairness, and robustness. 


\begin{figure}[hbt]
    \centering
    \includegraphics[width=\columnwidth]{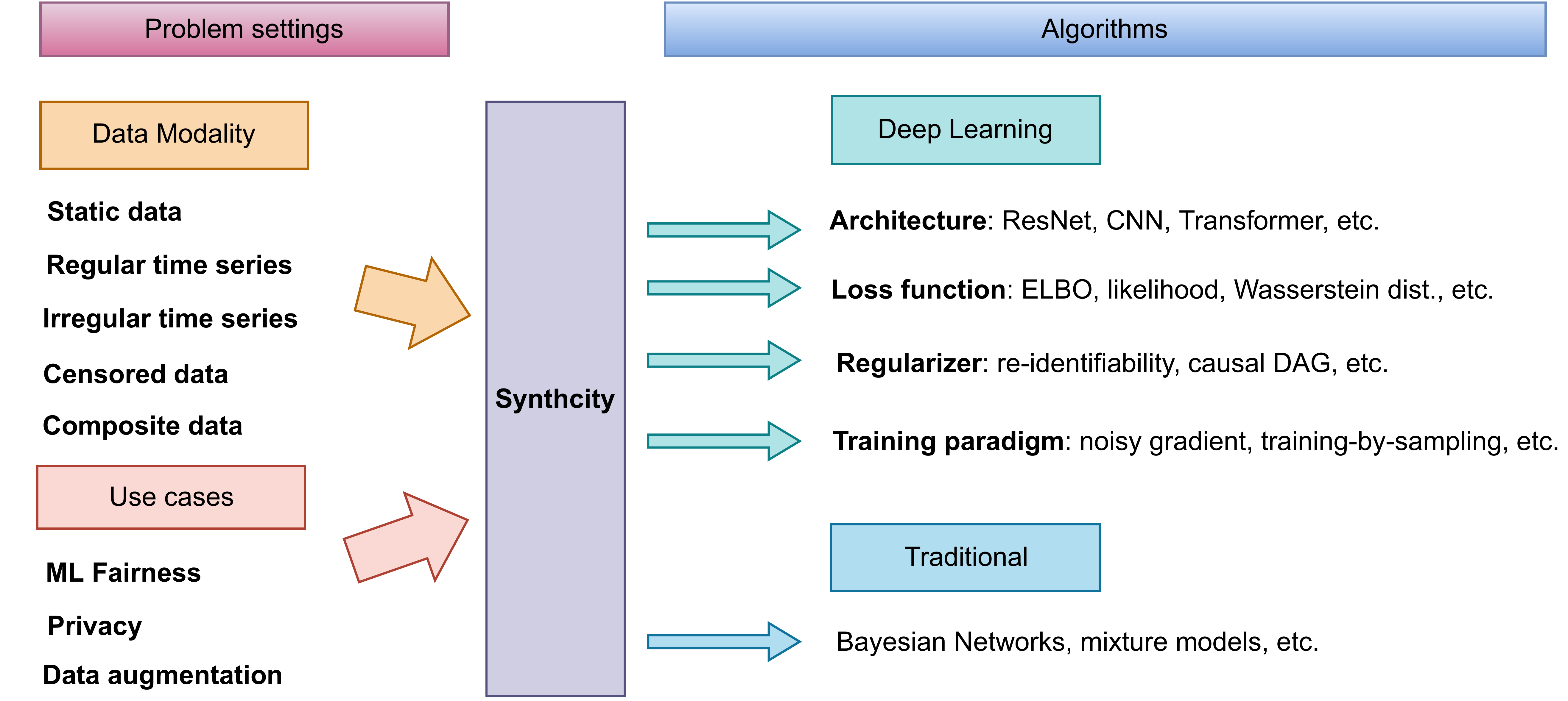}
    \caption{Synthcity covers diverse problem settings by mapping different data modalities and use cases to a host of deep learning and traditional data generation algorithms. }
    \label{fig:1}
\end{figure}

\section{The software challenge for synthetic data}

Despite the great progress in synthetic data research, its practical application is still in infancy. Much of the difficulty arises from two interlinked challenges in developing software for synthetic data generation, as depicted in Figure \ref{fig:1} and detailed below. 

\textit{1. The problem settings are diverse.}
The combination of different \textit{data modalities} and \textit{use cases} creates a large number of problem settings where no single generator (or class of generators) can expect to capture.
The datasets in practical applications have diverse modalities ranging from images, videos, and texts to tabular data which can be static, temporal or subject to various types of missingness and censoring. 
There also exist datasets collected from multiple sources, environments or views. 
Furthermore, the three main use cases (fairness, privacy and augmentation) involve different formalism, assumptions and methodology. 
Hence, while there exist many software libraries that can handle certain specific settings, a general platform that provides solutions to a wide range of problems is still lacking (as shown later in Table \ref{tab:compare}). 
This creates an obstacle for wide adoption of synthetic data technology.
For instance, a practitioner may very often find that the problem at hand cannot be solved by any of the existing software. 
In addition, having many specialized software also creates difficulty in software testing and maintenance, which are essential for high stake applications. The practitioner may also find it difficult to trust a library that has only been applied to narrow scenarios by a small user community.


\textit{2. The model choice is contextual.}
The community has developed a large number of generative models in the last decades. However, all models have their unique area of strengths and they encode different prior assumptions and inductive biases \citep{Bond-Taylor2021DeepModels}. Hence, it is no surprise that prior works have repeatedly shown that no model consistently performs the best across all data modalities and use cases.
For instance, deep generative models, such as GANs, excel at large image datasets, but they may fall short at small tabular data if not carefully trained and regularized.
This observation brings the conclusion that practitioners need a big arsenal of methods available at hand to address the diverse problem settings encountered in real applications. 
Yet, many cutting edge methods are currently implemented in individual code repositories that are not modular, reusable or interoperable.
Such repositories are usually made public along with the accepted paper and only implements one or few methods. Although beneficial for reproducibility, these repositories often cannot be easily extended, reused or combined. 
This creates a significant hurdle for experimenting and comparing different methods or extend them to broader settings. 
Creating a master repository that superficially collects many individual repositories is not an adequate solution because it does not fundamentally address the interoperability issue. 




In order to enable wider adoption of synthetic data and facilitate translational research in this promising area, the community needs a software platform that implements a large collection of state-of-the-art generators in a modular, reusable and composable way.


\section{The synthcity library}

To address the challenges above, we started the synthcity project, an open-source  initiative to create a software platform that facilitates innovative use cases of synthetic data in fairness, privacy and augmentation across diverse data modalities. 

We are now excited to announce the initial beta release of synthcity library (available on \href{https://pypi.org/project/synthcity/}{pip} and \href{https://github.com/vanderschaarlab/synthcity}{GitHub}). Synthcity is the first software library that covers all the major use cases of synthetic data including fairness, privacy, and augmentation in addition to the standard generation task. Furthermore, it also provides a diverse list of evaluation metrics to measure the quality of synthetic data. Last but not least, synthcity contains many utility functions to automate and streamline the workflow (e.g. automatically benchmarking the performance of multiple data generators).

In this version, we have focused on the generation of \textit{tabular data} due to its commonality in various industries and applications.
Tabular data is widespread in industries where data is based on relational databases including healthcare, finance, manufacturing etc. It is often in these regulated industries that data access is a challenge, making synthetic data beneficial.
We also emphasize that ``tabular data'' encapsulates many different data modalities, including static tabular data, time series data, and censored survival data, all of which may contain a mix of continuous and discrete features (columns). Synthcity can also handle composite datasets composed of multiple subsets of data. In future versions, we plan to include more data modalities and implements additional generators.
Synthcity is a live project under continuous development. We cordially invite the community to join the development effort by submitting comments, bug reports, and pull requests on GitHub.

\subsection{Overview of the synthcity workflow}

\begin{figure}[t!]
    \centering
    \includegraphics[width=\columnwidth]{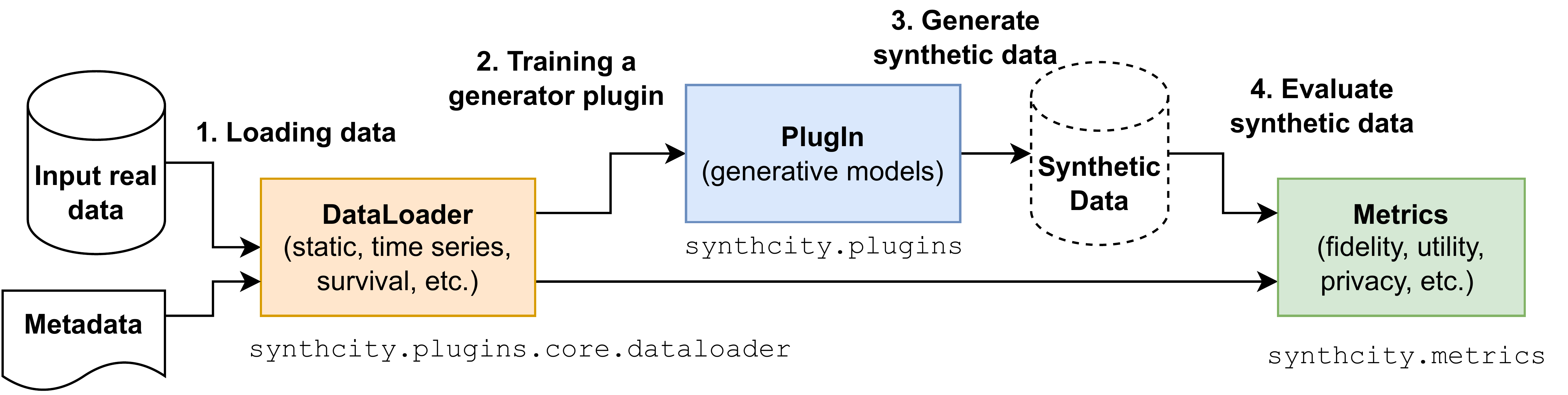}
    \caption{Standard workflow of generating and evaluating synthetic data with synthcity. }
    \label{fig:2}
\end{figure}

The synthcity library captures the entire workflow of synthetic data generation and evaluation. The typical workflow contains the following steps, as illustrated in Figure \ref{fig:2}:

\begin{enumerate}
    \item \textbf{Loading the dataset using a DataLoader}. The DataLoader class provides a consistent interface for loading and storing different types of input data (e.g. tabular, time series, and survival data). The user can also provide meta-data to inform downstream algorithms (e.g. specifying the sensitive columns for privacy-preserving algorithms).
    \item \textbf{Training the generator using a Plugin}. In synthcity, the users instantiate, train, and apply different data generators via the Plugin class. Each Plugin represents a specific data generation algorithm. The generator can be trained using the fit() method of a Plugin.
    \item \textbf{Generating synthetic data}. After the Plugin is trained, the user can use the generate() method to generate synthetic data. Some plugins also allow for conditional generation.
    \item \textbf{Evaluating synthetic data}. Synthcity provides a large set of metrics for evaluating the fidelity, utility, and privacy of synthetic data. The Metrics class allows users to perform evaluation. 
\end{enumerate}

In addition, synthcity also has a Benchmark class that wraps around all the four steps, which is helpful for comparing and evaluating different generators. 
After the synthetic data is evaluated, it can then be used in various downstream tasks.

\subsection{Data modalities}

\subsubsection{Single dataset}

We start by introducing the most fundamental case where there is a single training dataset (e.g. a single DataFrame in Pandas). We characterize the data modalities by two axes: the \textit{observation pattern} and the \textit{feature type}. 

The observation pattern describes whether and how the data are collected over time. There are three most prominent patterns, all supported by synthcity:

\begin{enumerate}
    \item \textit{Static}. All features are observed in a snapshot. There is no temporal ordering. 
    \item \textit{Regular time series}. Observations are made at regular intervals $t \in \{1, 2, \dots, T_i\}$, where $T_i \in \mathbb{N}^*$ is the maximum time horizon for series $i \in [N]$. Of note, this implies that different series may have different number of observations.  
    \item \textit{Irregular time series}. Observations are made at irregular intervals $(t_{i1}, t_{i2}, \dots, t_{iT_i})$, where  $T_i \in \mathbb{N}^*$, $\forall j \in  [ T_i ], t_{ij} \in \mathbb{R}^+$, and $i \in [N]$ represents different series. Note that, for different series $i$, the observation times may vary. 
\end{enumerate}

The feature type describes the domain of individual features. Synthcity supports the following three types. It also supports multivariate cases with a mixture of different feature types.

\begin{enumerate}
    \item Continuous feature, $x\in [a, b] \subset \mathbb{R}$, where $a,b \in \mathbb{R}$ are the lower and upper bounds.
    \item Categorical feature, $x\in \mathbb{S}$, where $\mathbb{S}$ is a finite set of allowed categories. 
    \item Integer feature, $x\in [a, b] \subset \mathbb{Z}$, where $a,b \in \mathbb{Z}$  are the lower and upper bounds.
    \item Censored feature, $(x, c)$, where $x\in \mathbb{R}^+$ represents the survival time and $c \in \{0, 1\}$ is the censoring indicator. 
\end{enumerate}

The combination of observation patterns and feature types give rise to an array of data modalities (as illustrated in Table \ref{tbl:modality}). Synthcity supports all combinations. 

\begin{table}[t!]
\centering
\begin{tabular}{@{}lcccc@{}}
\toprule
Obs pattern \textbackslash feature type & Continuous & Categorical& Integer & Censored \\ \midrule
Static                                          & $\surd$       & $\surd$  & $\surd$        & $\surd$     \\
Regular time series                             & $\surd$       & $\surd$  & $\surd$        & $\surd$     \\
Irregular time series                           & $\surd$       & $\surd$  & $\surd$        & $\surd$     \\ \bottomrule
\end{tabular}
\caption{synthcity supports combinations of different observation patterns and feature types.}
\label{tbl:modality}
\end{table}

\subsubsection{Composite dataset}

A composite dataset involves multiple sub datesets. For instance, it may contain datesets collected from different sources or domains (e.g. from different countries). It may also contain both static and time series data. Such composite data are quite often seen in practice. For example, a patient's medical record may contain both static demographic information and longitudinal follow up data. 

synthcity can handle the generation of differnet classes of composite datasets. Currently, it supports (1) multiple static datasets, (2) a static and a regular time series dataset, and (3) a static and a irregular time series dataset.

\subsubsection{Metadata}

Very often we have access to metadata that describes the properties of the underlying data. Synthcity can make use of these information to guide the generation and evaluation process. It supports the following types of metadata:

\begin{enumerate}
    \item sensitive features: indicator of sensitive features that should be protected for privacy
    \item outcome features: indicator of outcome feature that will be used as the  target in downstream prediction tasks. 
    \item domain: information about the data type and allowed value range. 
\end{enumerate}

\subsubsection{Missing data}

Currently, synthcity does not support training and generation of data with missingness. The user needs to first impute the missing data with an additional tool such as \href{https://pypi.org/project/hyperimpute/}{HyperImpute} 
 \citep{jarrett2022hyperimpute} and then present synthcity with the imputed data. 
We plan to add native support for missing data in future versions of synthcity. 

\subsection{Use cases and algorithms}

In this Section, we go through the use cases of synthetic data and the corresponding synthcity Plugins and algorithms. 

\begin{table}[t!]
\centering
\begin{tabular}{@{}llccccc@{}}
\toprule
\multirow{2}{*}{Data Modality} & \multirow{2}{*}{Plugin} & \multirow{2}{*}{Standard Gen} & \multicolumn{2}{c}{Privacy} & \multicolumn{2}{c}{Fairness} \\
 &  &  & DP & TM & Balance & Causal \\ \midrule
\multirow{10}{*}{Static} & Bayesian Net & $\surd$ &  &  &  &  \\
 & NF & $\surd$ &  &  &  &  \\
 & TVAE & $\surd$ &  &  & $\surd$ &  \\
 & RTVAE & $\surd$ &  &  & $\surd$ &  \\
 & CTGAN & $\surd$ &  &  & $\surd$ &  \\
 & PrivBayes & $\surd$ & $\surd$ &  &  &  \\
 & DPGAN & $\surd$ & $\surd$ &  &  &  \\
 & PATEGAN & $\surd$ & $\surd$ &  &  &  \\
 & ADSGAN & $\surd$ &  & $\surd$ &  &  \\
 & DECAF & $\surd$ &  &  &  & $\surd$ \\ \midrule
 \multirow{4}{*}{Static (Censored)} & Survival GAN & $\surd$ &  & $\surd$ & $\surd$  \\
 & Survival VAE & $\surd$ &  &  &  &  \\
 & Survival CTGAN & $\surd$ &  &  &  &    \\
 & Survival NF & $\surd$ &  &  &  &    \\ \midrule
 & TimeGAN & $\surd$ &  &  & $\surd$ &  \\
Time Series & TimeVAE & $\surd$ &  &  &  &  \\
(regular, irregular,  & FourierFlow* & $\surd$ &  &  &  &  \\
 censored) & Probabilistic AR* & $\surd$ &  &  &  &  \\ \midrule
Multi-source & RadialGAN & $\surd$ & \multicolumn{1}{l}{} & \multicolumn{1}{l}{} & $\surd$ & \multicolumn{1}{l}{} \\ \bottomrule
\end{tabular}%
\caption{Plugins available in synthcity for different data modalities and use cases. Abbreviations: Differential Privacy (DP), Threat Model (TM). *FourierFlow and Probabilistic AR  is compatible with regular time series only while TimeGAN and TimeVAE support both. }
\label{tab:ref_algorithm}
\end{table}

\begin{table}[t!]
\centering
\begin{tabular}{@{}ccc@{}}
\toprule
Static & Censored & Time Series \\ \midrule
Fully connected & Weibull AFT & LSTM \\
Residual network & Cox PH & GRU \\
 & Random Survival Forest & RNN \\
 & Survival Xgboost & Transformer \\
 & Deephit & MLSTM\_FCN \\
 & Tenn & TCN \\
 & Date & InceptionTime \\
 &  & InceptionTimePlus \\
 &  & XceptionTime \\
 &  & ResCNN \\
 &  & OmniScaleCNN \\
 &  & XCM \\ \bottomrule
\end{tabular}%
\caption{Available network architectures and survival models in synthcity for different data modalities. These components are compatible with multiple algorithms. }
\label{tab:architecture}
\end{table}

\subsubsection{Standard data generation}

Standard data generation refers to the most basic generation task, where the synthetic data should be generated as faithful as possible to the real-data distribution. Most existing works on generative modeling fall into this category \citep{Bond-Taylor2021DeepModels}. 

Formally speaking, let $X$ be the random variable of interest (which could be static, temporal or censored). The generator is trained using the training set $\mathcal{D} = \{x_i\}_{i=1}^N$, where $x_i \sim P(X)$ are sampled from the true (but unknown) distribution. During training, the generator (explicitly or implicitly) learns the distribution $\hat{P}(X)$ in order to sample from it. Typically, the training is performed by minimizing certain distributional distance between $P(X)$ and $\hat{P}(X)$, such as the Wasserstein distance and KL-divergence. 

Table \ref{tab:ref_algorithm} lists the plugins in synthcity for different data modalities. All of them support standard generation with appropriate configurations, although some may have additional use cases to be discussed later. 
Many algorithms above are based on deep neural networks. As such, the user can further specify their network architecture for the problem at hand (for instance, CNNs are better suited for highly frequently sampled time series than LSTM or Transformer). 
Table \ref{tab:architecture} lists the network architectures that are compatible with each data modality.

\subsubsection{Synthetic data for ML fairness}

Existing research have considered two different ways where Synthetic data could promote fairness. Table \ref{tab:ref_algorithm} shows the corresponding plugins in synthcity.

\textit{1. Balancing distribution}. In this setting, certain group(s) of people is underrepresented in the dataset for training downstream ML systems, which may lead to bias \citep{Lu2018contextual_bias,manela2021stereotype,kadambi2021achieving}. As a remedy, one could generate synthetic data of the minority group to augment the real data, thereby achieving balance in distribution. This often requires the data generator to learn the conditional distribution $P(X|G)$, where $G$ is the group label.

\textit{2. Causal fairness}. The second approach is to generate fairer synthetic data from a biased real dataset and to use synthetic data alone in downstream tasks \citep{zemel2013learning, xu2018fairgan,xu2019achieving,van2021decaf}. In this setting, it is postulated that the real distribution $P(X)$ reflects existing biases (e.g. unequal access to healthcare). The task for the generator is to learn a distribution $\hat{P}(X)$ that is free from such biases but also stay as close to $P(X)$ as possible (to ensure high data fidelity). Typically, notions of causality are employed in the bias removal process. This approach is also compatible with a host of criterion for algorithmic fairness, such as Fairness Through Unawareness, Demographic Parity and Conditional Fairness.

\subsubsection{Synthetic data for privacy}

Methods for generating privacy-preserving synthetic data mainly fall into two categories: the ones that employ differential privacy, and the ones that are designed to defend against specific attack modes.

\textit{1. Differential privacy (DP)}. DP is a formal way to describe how private a data generator is \citep{dwork2008differential}. Typically, generators with DP property introduces additional noise in the training procedure \citep{Jordon2022SyntheticHow}. For example, adding noise in the gradient or using a noisy discriminator in a GAN architecture \citep{Abadi2016DeepPrivacy, jordon2018pate, Long2019G-PATE:Discriminators}. It is worth noticing that DP is a formal property of the data generator, rather than the synthetic dataset. Hence, it is difficult to empirically verify if a particular synthetic data is private in a DP sense without knowing the exact data generating procedure. 

\textit{2. Threat model (TM)}. While DP focuses on giving formal guarantees, the TM approach is designed for specific threat models, such as membership inference, attribute inference, and re-identification \citep{shokri2017membership, kosinski2013private,dinur2003revealing}. The goodness of a TM-powered generator can be empirically evaluated using simulated attacks. However, such synthetic data are still subject to attack modes that are not considered or still unknown.

\subsubsection{Cross domain augmentation}

Here we consider a composite dataset that is collected from multiple domains or sources (e.g. data from different countries). Often one is interested in augmenting one particular data source that suffers from data scarcity issues (e.g. it is difficult to collect data from remote areas) by leveraging other related sources. 

This challenges has been studied in the deep generative model literature \citep{antoniou2017data, dina2022effect, das2022conditional, Bing2022ConditionalPopulations}. By learning domain-specific and domain-agnostic representations, the generator is able to transfer knowledge across domains, making data augmentation more efficient. Synthcity currently supports this mode of cross-domain generation. 

\subsection{Evaluation of synthetic data}

Before releasing or using synthetic data in any downstream task, the quality of the data needs to be evaluated first. Synthetic data evaluation is also important when it comes to the selection and comparison of generative models. Synthcity provides the users with a comprehensive list of  metrics to evaluate various aspects of synthetic data. The metrics broadly measures three aspects of synthetic data (outlined below). A full list of metrics can be found in Table \ref{tab:metric}. 

\textit{Fidelity}. The fidelity of synthetic data captures how much the synthetic data resembles real data. The fidelity metrics typically evaluate the closeness between the true distribution $P$ and the distribution learned by the generator $\hat{P}$ using samples from these two distributions. Synthcity supports distributional divergence measures (e.g. Jensen-Shannon distance, Wasserstein distance, and maximal mean discrepancy) as well as two sample detection scores (i.e. using a classifier to distinguish real verses synthetic data) \citep{snoke2018general}. 

\textit{Utility}. The utility of synthetic data reflects how useful the synthetic data is to a downstream task. This captures the common scenario of train-on-synthetic evaluate-on-real \citep{beaulieu2019privacy}. Synthcity supports various types of downstream tasks, including regression, classification and survival analysis. In addition to linear models, synthcity supports xgboost and neural nets as downstream  models due to their wide adoption in analytics. 

\textit{Privacy}. Synthcity includes a list of well-established privacy metrics (e.g. k-anonymity and l-diversity). Furthermore, it can measure privacy of data by performing simulated privacy attack (e.g. re-identification attack). The success (or failure) of such attack quantifies the degree of privacy preservation.

\section{Comparison with existing libraries}

In this section, we compare synthcity with other popular open source libraries for synthetic data. Table \ref{tab:compare} shows that synthcity supports much broader data modalities and use cases than the alternatives. A more detailed comparison of the supported data generators and evaluation metrics are available in Table \ref{tab:ref_algorithm} and \ref{tab:metric}.

\begin{table}[t!]
\centering
\resizebox{\textwidth}{!}{%
\begin{tabular}{@{}lccccccc@{}}
\toprule
Setting \textbackslash Software & Synthcity & YData & Gretel & SDV & DataSynthesizer & SmartNoise & nbsynthetic \\ \midrule
\textbf{Data modalities} &  &  &  &  &  &  &  \\
Static data & $\surd$ & $\surd$ & $\surd$ & $\surd$ & $\surd$ & $\surd$ & $\surd$ \\
Regular time series & $\surd$ & $\surd$ & $\surd$ & $\surd$ &  &  &  \\
Irregular time series & $\surd$ &  &  &  &  &  &  \\
Censored features & $\surd$ &  &  &  &  &  &  \\
Composite data & $\surd$ &  &  & $\surd$ &  &  &  \\ \midrule
\textbf{Use cases} &  &  &  &  &  &  &  \\
Generation & $\surd$ & $\surd$ & $\surd$ & $\surd$ & $\surd$ & $\surd$ & $\surd$ \\
Fairness (balance) & $\surd$ & $\surd$ & $\surd$ & $\surd$ &  &  & $\surd$ \\
Fairness (causal) & $\surd$ &  &  &  &  &  &  \\
Privacy (DP) & $\surd$ &  & $\surd$ &  &  & $\surd$ &  \\
Privacy (TM) & $\surd$ &  &  &  &  &  &  \\
Cross domain aug. & $\surd$ &  &  &  &  &  &  \\ \bottomrule
\end{tabular}%
}
\caption{The data modalities and use cases supported by synthcity and other open source synthetic data libraries. Comparisons are based on the software versions available at the time of writing. }
\label{tab:compare}
\end{table}

\begin{table}[ht]
\centering
\resizebox{\textwidth}{!}{%
\begin{tabular}{@{}llccccccc@{}}
\toprule
Aspect & Evaluation Metric \textbackslash Software & Synthcity & YData & Gretel & SDV & DataSynthesizer & SmartNoise & nbsynthetic \\ \midrule
\multirow{15}{*}{Fedelity} & Jensen-Shannon distance & $\surd$ &  &  &  &  &  &  \\
 & Wasserstein distance & $\surd$ &  &  &  &  &  &  \\
 & Total variation distance &  &  &  & $\surd$ &  &  &  \\
 & KL divergence & $\surd$ &  &  &  &  &  &  \\
 & Skewness &  &  &  & $\surd$ &  &  &  \\
 & Max-mean discrepancy & $\surd$ &  &  &  &  &  & $\surd$ \\
 & KS test & $\surd$ &  &  & $\surd$ &  &  & $\surd$ \\
 & PRDC & $\surd$ &  &  &  &  &  &  \\
 & Alpha--precision & $\surd$ &  &  &  &  &  &  \\
 & Survival Kaplan-Meier dist. & $\surd$ &  &  &  &  &  &  \\
 & Detection: linear & $\surd$ &  &  & $\surd$ &  &  &  \\
 & Detection: NN & $\surd$ &  &  &  &  &  &  \\
 & Detection: XGB & $\surd$ &  &  &  &  &  &  \\
 & Detection: GMM & $\surd$ &  &  & $\surd$ &  &  &  \\
 & Detection: Bayesian &  &  &  & $\surd$ &  &  &  \\ \midrule
\multirow{6}{*}{Utility} & Linear model & $\surd$ &  &  & $\surd$ &  &  &  \\
 & MLP & $\surd$ &  &  & $\surd$ &  &  &  \\
 & XGBoost & $\surd$ &  &  & $\surd$ &  &  &  \\
 & Static survival & $\surd$ &  &  &  &  &  &  \\
 & Time-series & $\surd$ &  &  &  &  &  &  \\
 & Survival time-series & $\surd$ &  &  &  &  &  &  \\ \midrule
\multirow{6}{*}{Privacy} & Correct attribution prob. & $\surd$ &  &  & $\surd$ &  &  &  \\
 & K-anonymity & $\surd$ &  &  &  &  &  &  \\
 & K-map & $\surd$ &  &  &  &  &  &  \\
 & Delta-presence & $\surd$ &  &  &  &  &  &  \\
 & L-diversity & $\surd$ &  &  &  &  &  &  \\
 & Identifiability score & $\surd$ &  &  &  &  &  &  \\ \bottomrule
\end{tabular}%
}
\caption{The evaluation metrics supported by synthcity and other open source synthetic data libraries. Comparisons are based on the software versions available at the time of writing.}
\label{tab:metric}
\end{table}

\section{Example usage scenarios}

To better contextualize the utility of synthcity, here we conceive several illustrative use cases in different industries. This is by no means a comprehensive list, and we encourage the reader to come up with other usage scenarios and share the story with us.  


\subsection{De-biasing medical datasets to build better and fairer prognostic scores}

Electronic health records (EHR) and biobanks can be used by medical researchers to build disease prognostic scores. Such scores allow the healthcare service to allocate medical resources more efficiently (e.g. performing triage). They can also inform the design and execution of clinical trials (e.g. stratifying patient population). 

However, EHR and biobanks may often under-represent certain populations (e.g. the ones with less privileged access to health service). As a result, naively building prognostic scores on these data may lead to undesirable and unfair decisions. To mitigate the bias, one may re-balance the real data by augmenting it with synthetically generated minority groups. Synthcity facilitates this use case by providing a variety of standard, conditional, and multi-domain generative models as well as tools for evaluating the synthetic data. 



\subsection{Using synthetic multi-source data to select the target audience of marketing campaigns}

To maximize the performance of a digital marketing campaign, the marketeer needs to locate the right customer segment to display the ad. This requires analyzing historical data from different segments. However, the data might be scarce for certain customer segments (e.g. the VIP users) and cannot fully support the analytics. In this case, one may consider augmenting the data using knowledge from other segments. Synthcity supports cross-domain data augmentation with deep generative models. The augmented synthetic data can then be combined with real data to inform the campaign setup.

\subsection{Outsourcing analytics to contractors or the open science community
}

The data holder may be interested in outsourcing the analytical modeling task to a contractor or the open science community (e.g. running a data science competition). However, the data holder cannot share real data with these external parties due to the sensitivity of data. A possible solution is to use the privacy-preserving generators in synthcity to create synthetic datasets. After evaluating the datasets in terms of fidelity and privacy, the data holder may share the synthetic data to the external parties to build the analytical models. The trained models can then be transferred to the data holders for evaluation and deployment on real data.

\section{Further involvement in the synthcity project}

We welcome the practitioners, researchers, and open source community to further engage in the synthcity project. 
We will run a series of tutorials and labs in the coming months to demonstrate the practical utility of synthcity.
If you would like to stay tuned for the upcoming events and releases, please sign up for our \href{https://forms.gle/rbXnwDUN8zonC8eR8}{mailing list}.

We are keen to receive your comments and feedback. Together, we can build a software to better extract the potentials of synthetic data technology.

\begin{table}[t!]
\centering
\resizebox{\textwidth}{!}{%
\begin{tabular}{@{}lccccccc@{}}
\toprule
Algorithm   \textbackslash Software & Synthcity & YData & Gretel & SDV & DataSynthesizer & SmartNoise & nbsynthetic \\ \midrule
CTGAN & $\surd$ & $\surd$ &  & $\surd$ &  &  & $\surd$ \\
ACTGAN &  &  & $\surd$ &  &  &  &  \\
TVAE & $\surd$ &  &  & $\surd$ &  &  &  \\
Bayesian Network & $\surd$ &  &  &  &  &  &  \\
Normalizing Flows & $\surd$ &  &  &  &  &  &  \\
Survial GAN & $\surd$ &  &  &  &  &  &  \\
Survival VAE & $\surd$ &  &  &  &  &  &  \\
DoppelGANger &  &  & $\surd$ &  &  &  &  \\
TimeGAN & $\surd$ & $\surd$ &  &  &  &  &  \\
FourierFlows & $\surd$ &  &  &  &  &  &  \\
Probabilistic AR & $\surd$ &  &  & $\surd$ &  &  &  \\
DECAF & $\surd$ &  &  &  &  &  &  \\
RadialGAN & $\surd$ &  &  &  &  &  &  \\
ADSGAN & $\surd$ &  &  &  &  &  &  \\
DPGAN & $\surd$ &  & $\surd$ &  &  & $\surd$ &  \\
PATEGAN & $\surd$ &  &  &  &  & $\surd$ &  \\
PrivBayes & $\surd$ &  &  &  & $\surd$ &  &  \\ \bottomrule
\end{tabular}%
}
\caption{The data generating algorithms supported by synthcity and other open source synthetic data libraries. Comparisons are based on the software versions available at the time of writing.}
\label{tab:algo-compare}
\end{table}

\newpage
\bibliography{ref}

\end{document}